\documentclass[conference]{IEEEtran}

\usepackage{cite}
\usepackage{amsmath,amssymb,amsfonts}
\usepackage{algorithmic}
\usepackage{graphicx}
\usepackage{textcomp}
\usepackage{xcolor}
\usepackage{fancyhdr}
\def\BibTeX{{\rm B\kern-.05em{\sc i\kern-.025em b}\kern-.08em
    T\kern-.1667em\lower.7ex\hbox{E}\kern-.125emX}}
\usepackage[none]{hyphenat}
\usepackage{hyperref}
\begin{document}

\title{%
  Complying with the EU AI Act\\
  \large On which areas should organizations focus when considering compliance with the AIA? \\
}

\author{\IEEEauthorblockN{Jacintha Walters, Diptish Dey, Debarati Bhaumik, Sophie Horsman}
\IEEEauthorblockA{
Amsterdam University of Applied Sciences \\
Amsterdam, 14-07-2023\\
jacintha.walters@hva.nl\\
}
}
\maketitle

\pagestyle{fancy}
\fancyhf{}
\cfoot{\thepage}
\fancyhead[L]{\leftmark}
\renewcommand{\footrulewidth}{0.3pt}
\rfoot{\includegraphics[width=4cm]{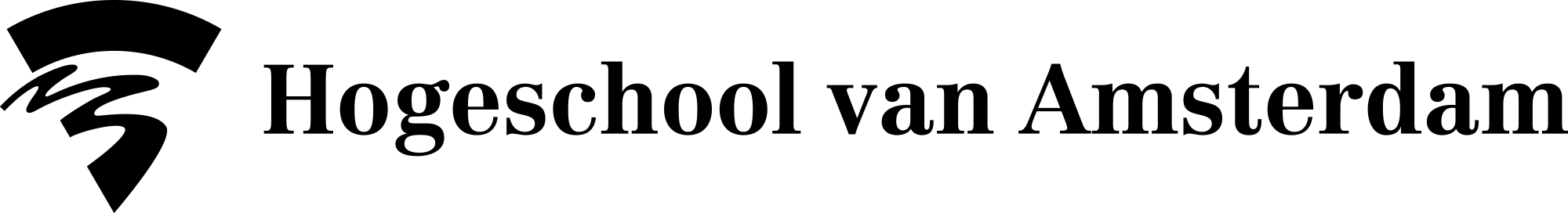}}

\begin{abstract}
The EU AI Act is the proposed EU legislation concerning AI systems. This paper identifies several categories of the AI Act. Based on this categorization, a questionnaire is developed that serves as a tool to offer insights by creating quantitative data. Analysis of the data shows various challenges for organizations in different compliance categories. The influence of organization characteristics, such as size and sector, is examined to determine the impact on compliance. The paper will also share qualitative data on which questions were prevalent among respondents, both on the content of the AI Act as the application. The paper concludes by stating that there is still room for improvement in terms of compliance with the AIA and refers to a related project that examines a solution to help these organizations.
\end{abstract}

\section{Introduction}
The EU AI Act (in this paper abbreviated as AIA) is a proposed regulation (law) by the European Commission that aims to regulate the application of artificial intelligence in the European Union. The proposed regulation was published in 2021 and is currently under review by the European Parliament and the Council of the European Union. It defines which AI systems are categorized as high-risk and the rules applicable before a high-risk AI can be used \cite{theact}. At the time of writing, it is unknown when the AIA will become in effect\cite{commission}.

One of the challenges in complying with the AIA is that AI systems are developed and maintained by a chain of actors, including software developers, data scientists, and engineers. The challenges for organizations are further complicated because of the interdisciplinary character of legal, technical, and domain-specific responsibilities. For an organization to comply, it must be able to interpret the contours implied by the act and translate this information into relevant requirements. \\

This paper identifies areas where organizations face challenges when considering current and future compliance with the AIA. The following steps are undertaken to identify these areas. Initially, categories of concern within the AIA are identified. Based on this categorization, a questionnaire is constructed to gather insights into how organizations handle the requirements associated with each category. The questionnaire is further refined through expert reflection, and trial runs to ensure its effectiveness.

Interactive interviews are conducted with multiple respondents based on the questionnaire. An online questionnaire is also circulated to obtain a broader range of responses. The obtained responses are then assessed using a three-point rating system. This evaluation process results in the computation of a "compliance score" for each category.

The next step involves analyzing the responses to identify focus areas where organizations encounter the most difficulties. Furthermore, the influence of organization characteristics, such as size and sector, is examined to determine their impact on compliance. Finally, by analyzing the gathered responses, the most prevalent questions among the respondents are identified, shedding light on the specific concerns organizations face regarding the AIA.

To ensure a manageable questionnaire size, a decision was made to exclude certain subjects discussed in the AIA. The subjects of robustness, cybersecurity, logging, reporting, and audit preparedness were omitted from the questionnaire. This exclusion was primarily driven by the need to reduce the questionnaire's length, making it more feasible for potential respondents to complete. Although relevant to AI development, these subjects are broader and primarily associated with IT development.

\section{Relevant Literature}
Usman {\it et al.} observes that organizations can be subject to multiple regulations, which may lead to several challenges. First, there can be conflicting requirements. Second, some regulations are not well-defined, leaving the development team unsure how to implement them. After implementation, it can be challenging to verify that the software system meets all the requirements \cite{ericsson}.

Research on privacy regulations showed that many small medium enterprises (SMEs) do not possess sufficient knowledge of the regulations to achieve compliance. Besides the risk of fines, compliance is essential to sustain if organizations want to supply services to other compliant organizations \cite{2018}.

Government-owned organizations generally demonstrate a lower degree of compliance than other organizations. Besides, there can be a significant difference in compliance for different sectors \cite{chua}.

Most existing research on the AIA has a theoretical perspective, focusing mainly on the quality of the content of the AIA rather than the application. One study concludes that the AIA is a good attempt but has several weaknesses. For instance, many parts are ambiguous, making it hard for organizations to define rules to self-assess against \cite{veale}. Another study concludes that the AIA is generally well-constructed but advises that the proposal should not rely so heavily on internal controls. External oversight is a necessity \cite{rail}. 

A notable research gap exists regarding the future compliance of organizations with the AIA and their level of preparedness. The existing literature covers two parts. First, compliance with existing regulations like GDPR. Second, critical analysis of the content of the AIA. There is a lack of insight into how organizations will navigate compliance with the AIA and the extent to which they are prepared. This paper aims to address this research gap by providing insights into the level of preparedness and the challenges organizations will face in complying with the AIA. 

\section{Methodology}
\subsection{Identifying Categories in the AIA}
\begin{figure}[ht]
    \centering
    \includegraphics[width=0.9\linewidth]{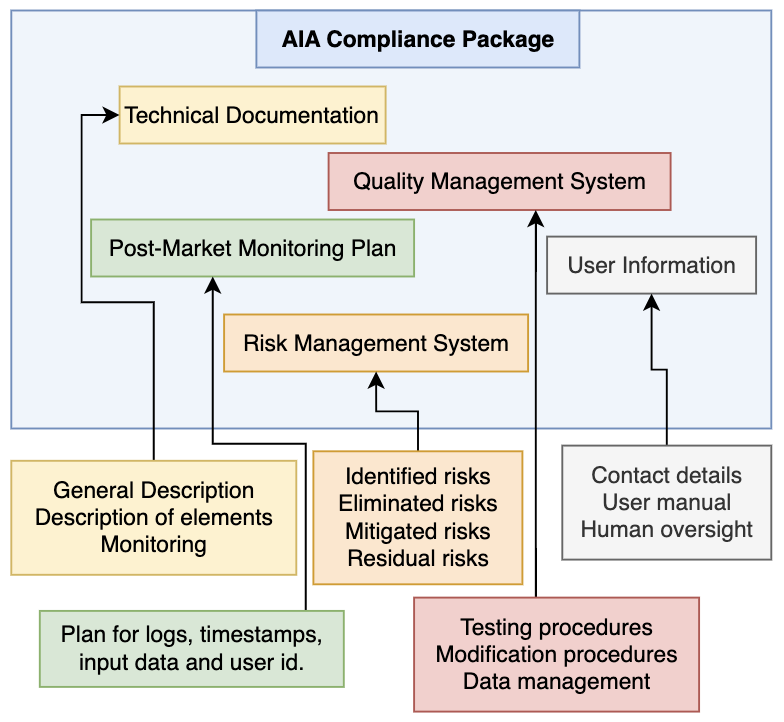}
    \caption{Overview of the AIA compliance documentation}
    \label{fig:aia}
\end{figure}

\begin{figure}[ht]
    \centering
    \includegraphics[width=0.8\linewidth]{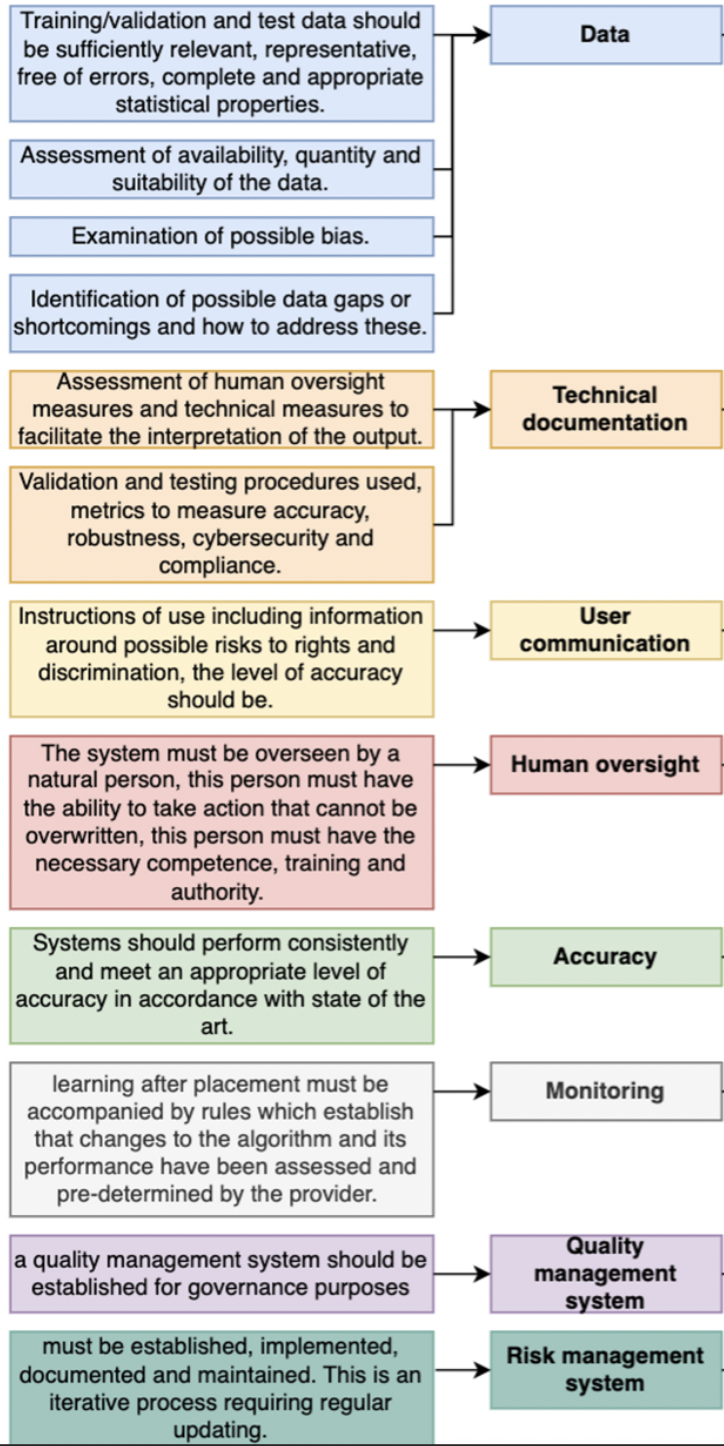}
    \caption{Overview of key subject areas from the AIA}
    \label{fig:funcreq}
\end{figure}

 Figure \ref{fig:aia} shows an overview of the relevant documentation for AIA compliance. Figure \ref{fig:funcreq} shows a hierarchical breakdown of key subject areas from the AIA used as a basis for the questionnaire.

\subsection{Creating and Refining the Questionnaire}
\begin{figure}[h]
    \centering
    \includegraphics[width=1.0\linewidth]{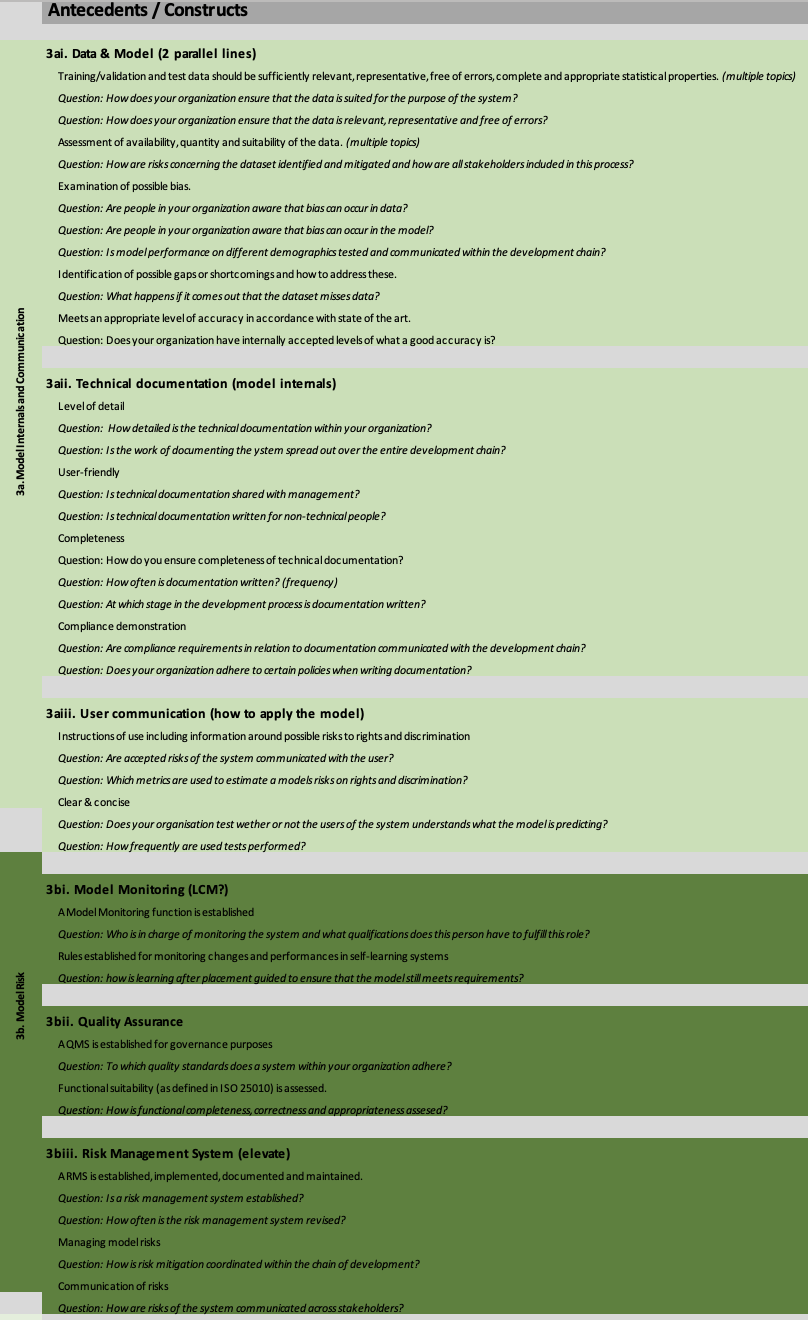}
    \caption{The conceptual framework}
    \label{fig:fram}
\end{figure}

 The hierarchical breakdown from Figure \ref{fig:funcreq} is used to create a conceptual framework (Figure \ref{fig:fram}). Throughout the construction of the conceptual framework, iterative feedback loops are established with domain experts to refine the framework's clarity and coherence. The final framework contains two parts. The first part focuses on model internals and communication. This part includes the data and model, the technical documentation, and the user communication. The second part focuses on model risk, including model monitoring, quality assurance, and risk management systems.

 The conceptual framework was expanded with open questions. These questions are set up through discussion with domain experts. The questions are in Figure \ref{fig:fram}.

 The construction of the questionnaire involves several iterative steps to improve its validity and reliability. Existing questions were rephrased to transform open-ended questions into closed questions. Proxy questions are incorporated to ensure fair and reasonable responses. For example, the statement ``My organization identifies and mitigates risks associated with a dataset'' is supported by the question, ``How often does your organization mitigate risks in a dataset?''. The questionnaire contains around 90 questions, which took respondents about 15 minutes to answer.\footnote{Questionnaire: https://hva.eu.qualtrics.com/jfe/form/SV\_9sFXWLoj5uFoaua} Feedback was gathered on the questionnaire from two organizations through an online interactive trial run. 

\begin{figure}[ht]
    \centering
    \includegraphics[width=1.0\linewidth]{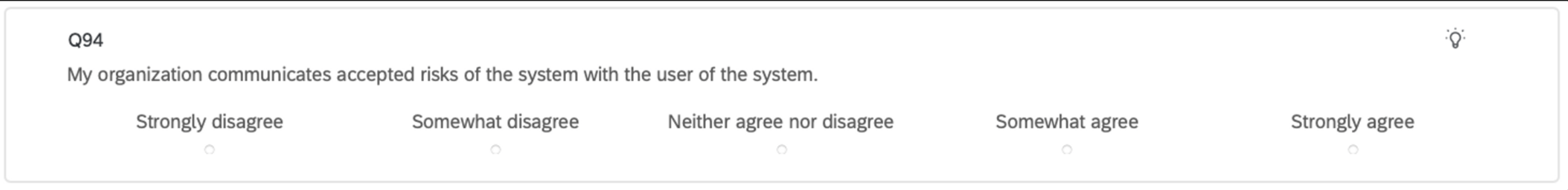}
    \caption{'Statement...' questions}
    \label{fig:state}
\end{figure}

\begin{figure}[ht]
    \centering
    \includegraphics[width=0.7\linewidth]{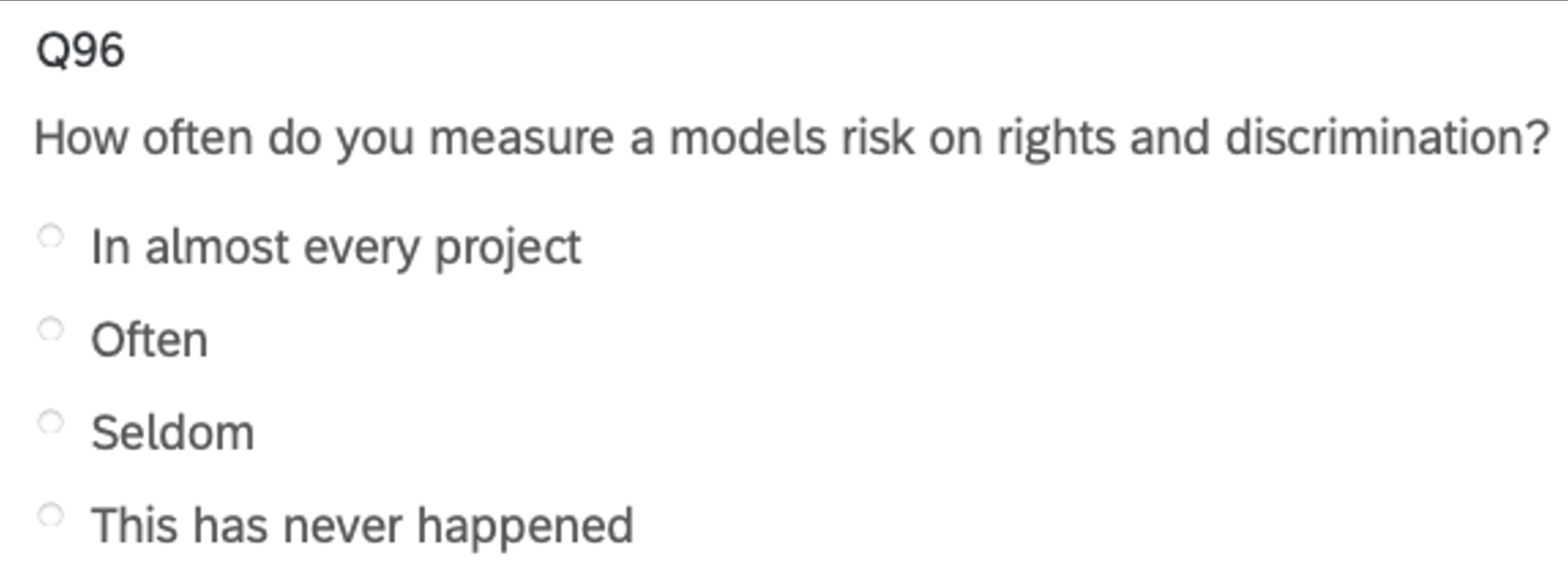}
    \caption{'How often...' questions}
    \label{fig:how}
\end{figure}

\begin{figure}[ht]
    \centering
    \includegraphics[width=0.7\linewidth]{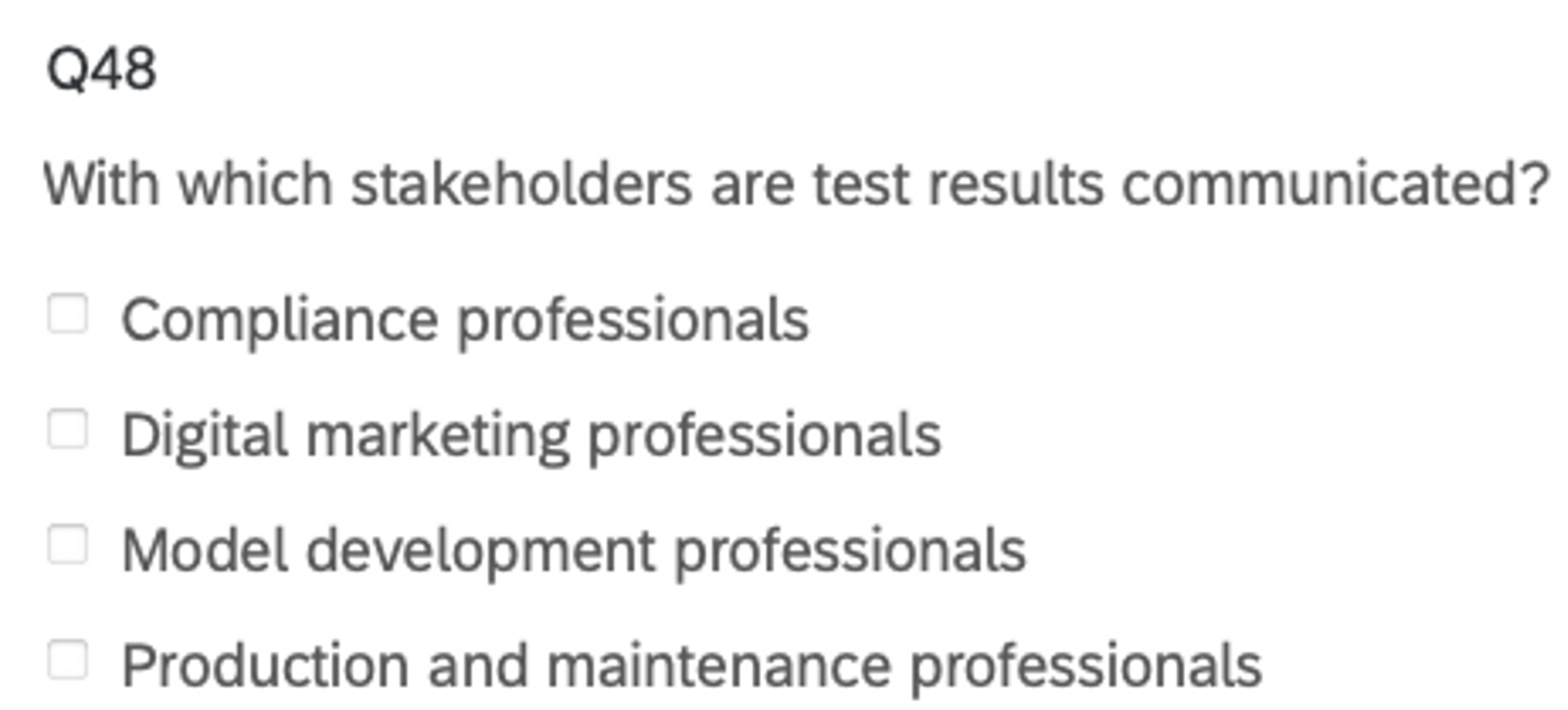}
    \caption{'Who...' questions}
    \label{fig:who}
\end{figure}

 Most questions follow one of three categories shown in Figures \ref{fig:state}, \ref{fig:how}, and \ref{fig:who}. ``Statement questions'' rely primarily on respondents' perspectives, whereas other questions are more objective. This observation is used to compute a ``compliance score''.


 Following the finalization of the questionnaire, a series of interactive interviews were conducted to collect responses. Additionally, the questionnaire was shared online, resulting in more responses. In total, seven responses were obtained through interactive interviews, supplemented by eight responses gathered online.

\subsection{Response Rating}
 Each of the fifteen responses is rated using a three-point range. The scoring process aims to quantify the responses for each questionnaire category to enable numeric comparison. 

 The questionnaire data is rated using a rule-based system. The rule-based system involves manually creating rules that are used to score each entry in the dataset automatically.\footnote{Code: https://gitfront.io/r/user-7646844/ZTQB4rfx5SYN/CustomLLM/} 

\begin{figure}[ht]
    \centering
    \includegraphics[width=1.0\linewidth]{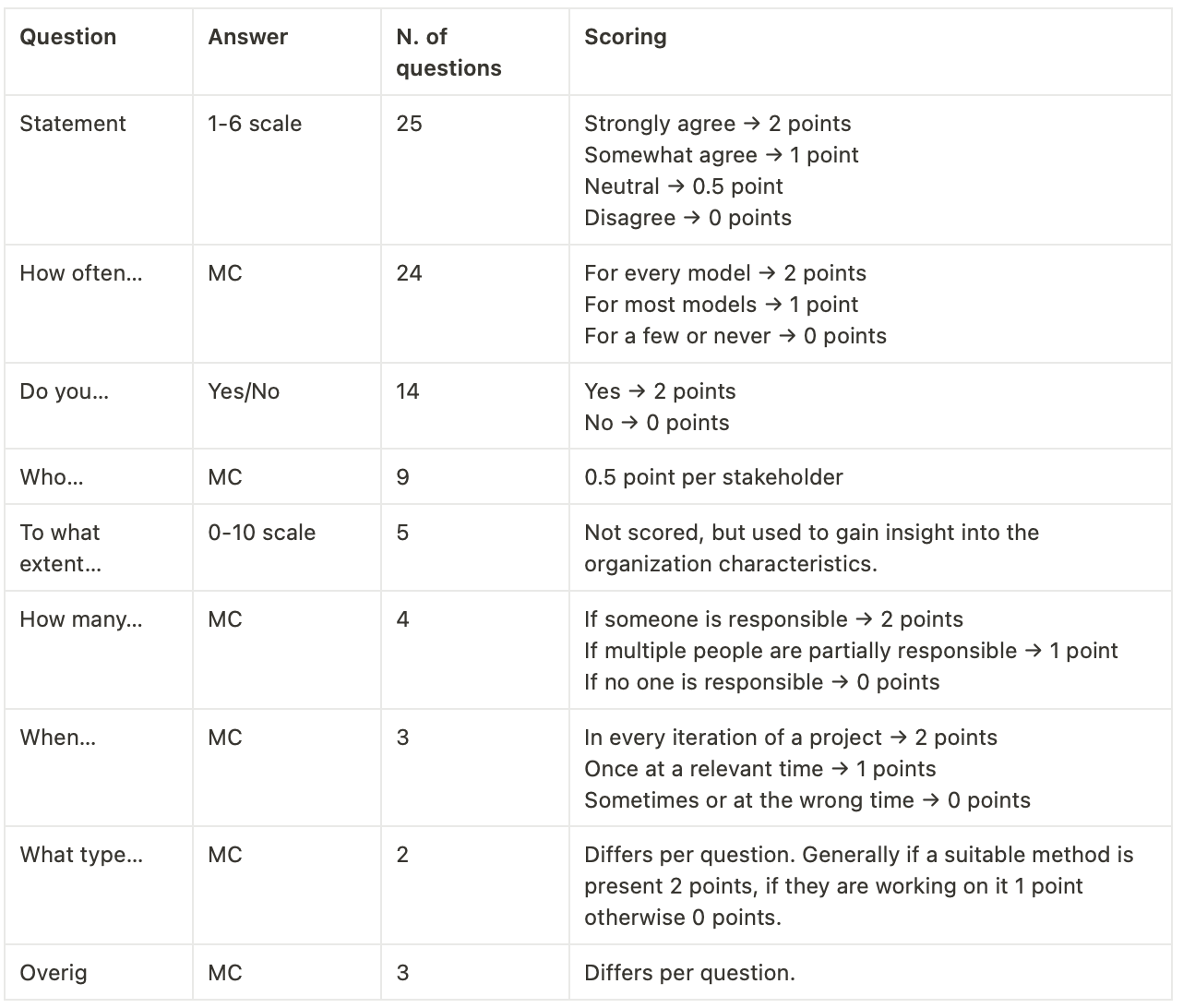}
    \caption{Points given per question type}
    \label{fig:pointsper}
\end{figure}

 The following categories from the questionnaire are used: data and model internals, technical documentation, user communication, model monitoring, and risk management. Generally, each question has a ``perfect'' answer worth 2 points, followed by ``reasonable'' answers worth 1 point. The remaining answers score 0 points. The perfect answer aligns closely with the requirements stated in the AI Act. The point distribution for each question is summarized in Figure \ref{fig:pointsper}.

\begin{figure}[ht]
    \centering
    \includegraphics[width=1.0\linewidth]{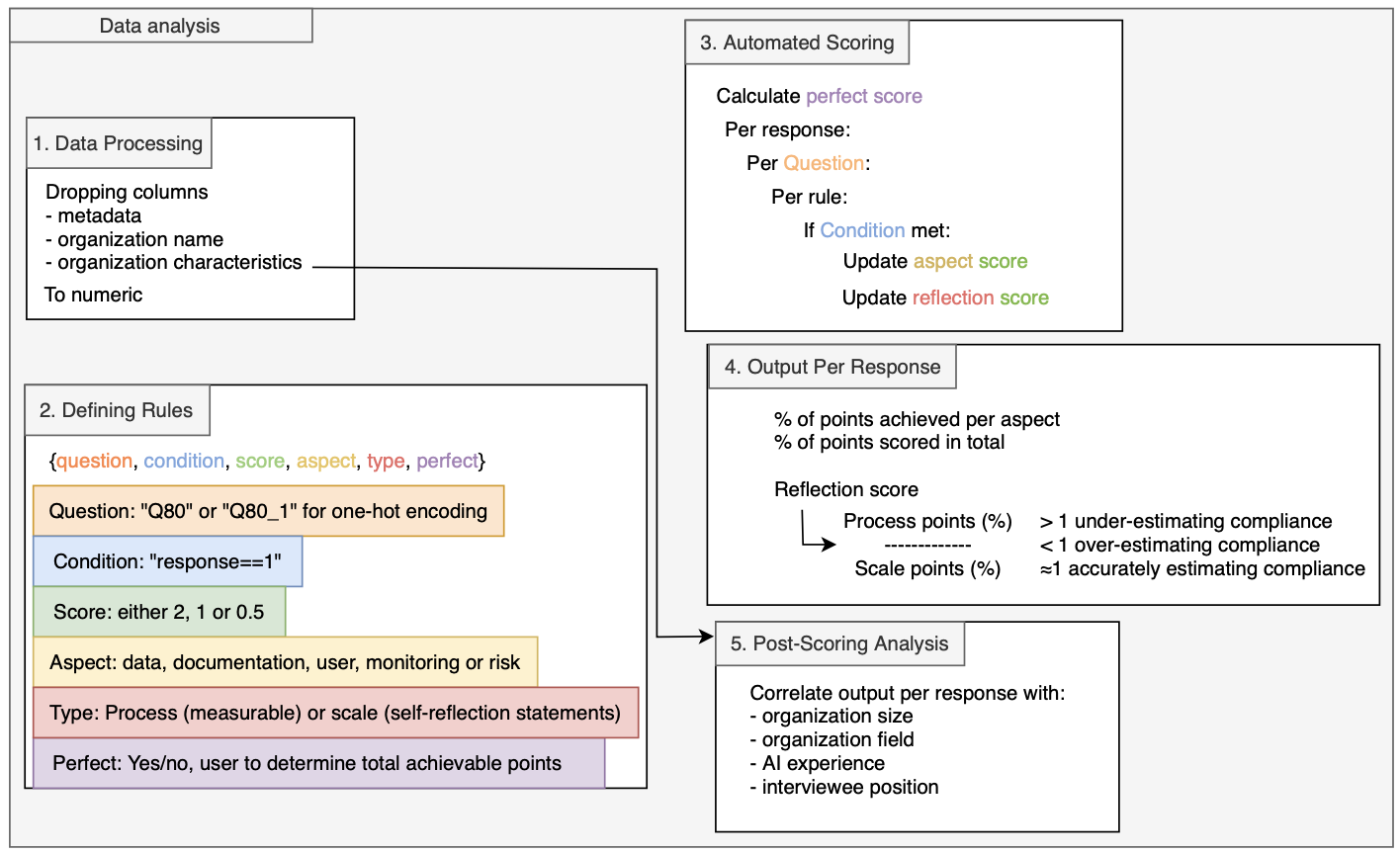}
    \caption{Scoring system of questionnaire data}
    \label{fig:autom}
\end{figure}

 The overview of how the automated scoring process was implemented is shown in Figure \ref{fig:autom}. Each respondent's score for each category is calculated along with the reflection score. The reflection score was added to show the validity of the responses. The reflection score is based on the ratio of ``statement questions'' and other questions (process). The reflection score determines if an organization over- or under-estimates itself.

\section{Results}
\subsection{Identifying Focus Areas}
\begin{figure}[ht]
    \centering
    \includegraphics[width=0.8\linewidth]{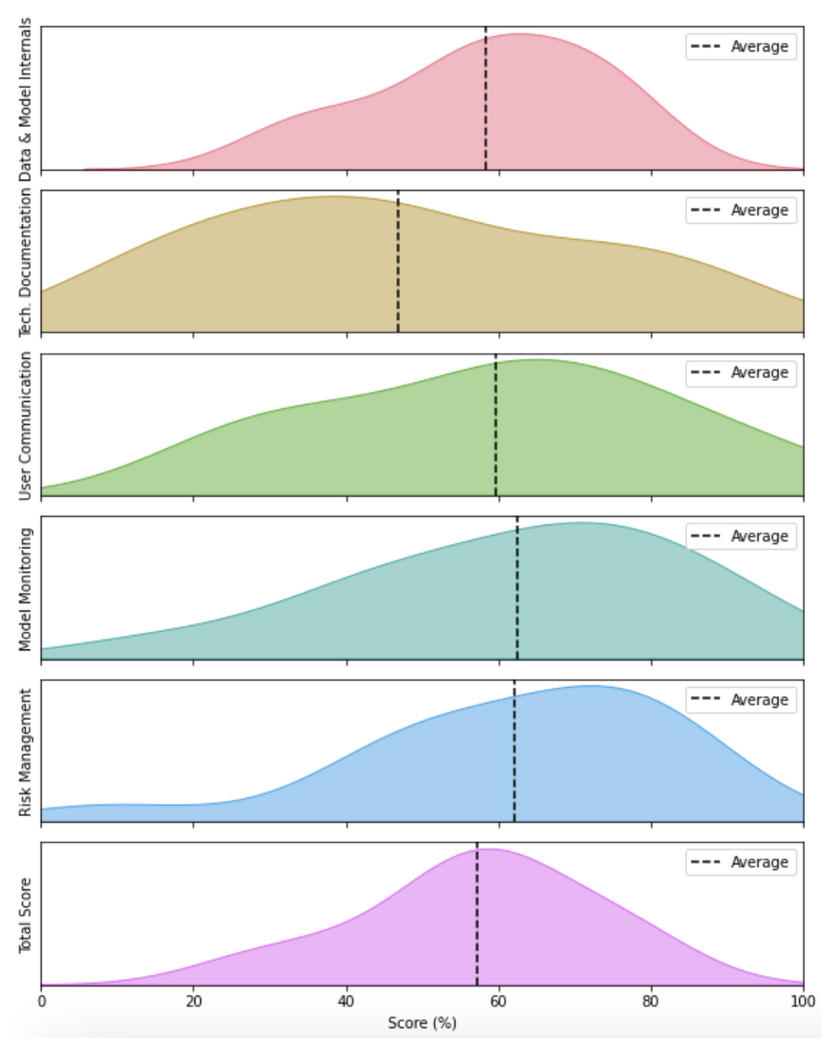}
    \caption{Percentage of points per questionnaire category}
    \label{fig:aspects}
\end{figure}

 The average percentage score for each category of the questionnaire is shown in Figure \ref{fig:aspects}. 

 The overall average compliance score for all respondents and categories is 57\%. The average reflection score is 1.0, suggesting that organizations demonstrate good self-awareness.

 Figure \ref{fig:aspects} reveals variations in compliance scores across different categories. The questionnaire results show that many organizations lack procedures for technical documentation and do not have someone trained to determine compliance requirements. Regarding data and model internals, organizations struggle with training employees on data and model bias. User communication presents challenges in determining metrics for measuring model risks on rights and discrimination. Risk management systems are found to be lacking in some organizations. Model monitoring shows a mixed trend, with some organizations adequately updating models when needed. Besides these challenges, the interactive interviews also identified several prevalent questions among respondents, both on the content of the AIA as the application within their organization. These questions are listed in Appendix A.

\subsection{Organization Characteristics' Influence}
 Organizations with 1-50 employees scored lower than organizations with 51+ employees. As for the industry, the dataset is too small to draw any conclusions. There is a large variance in compliance scores for the IT sector, from 26\% to 67\%. One organization felt that their ISO certification helped them to comply with the AIA. Organizations with more AI experience in years did not score better compared to organizations relatively new to AI. 

\section{Conclusion}
 This paper examines in which areas organizations seem to be struggling when considering current and future compliance with the AIA. A conceptual framework has been constructed based on a review of the act. A questionnaire is formulated based on the framework. Fifteen organizations answered the entire questionnaire.

 A compliance score is calculated using a rule-based system that awards points for answers following the contents of the AIA. Organizations achieve an average compliance score of 57\% compared to the perfect score. This score indicates there is room for improvement toward AIA readiness. Organizations are best prepared on model monitoring and risk management but score the lowest, with 47\%, on technical documentation.

 The duration of AI usage by organizations does not result in a higher compliance score. The same goes for IT organizations compared to non-IT organizations. 

 Overall, this paper contributes to the growing body of knowledge on the implementation of the AIA. The paper is the first to identify focus areas for different categories of the AIA to help organizations better prepare. Organizations will need help dealing with the questions and challenges they are facing. A custom large language model has been created in a related project to help organizations answer questions about the AIA \cite{babelfish}.

\section{Future Research}
 Several approaches for future research are identified. First, the predictive power of the questionnaire should be tested to see if the questionnaire can predict if an organization will pass the self-assessment. More qualitative data should be gathered by observing the AIA compliance processes. This data can then be used to refine the questionnaire for different organizations' sizes and sectors.

 \section{Acknowledgement}
 I want to acknowledge that this research was conducted as part of my master thesis for the Master Applied AI program at Amsterdam University of Applied Sciences (HvA). I am grateful for the guidance and support provided by Diptish Dey, Debarati Bhaumik, and Sophie Horsman, who served as my advisors throughout the research process. I would also like to thank the Centre for Market Insights (a research lab within the HvA) for their assistance and resources in facilitating this study.

\begin{appendices}
\section{Prevalent Questions}
Questions on the content of the AIA:
\begin{enumerate}
    \item Should technical documentation also be written for non-technical people?
    \item Does the AIA stipulate that we need someone to monitor the AI models full-time?
    \item Does the AIA require me to work with encrypted data only?
    \item How should we deal with missing data according to the AIA?
    \item What other data risks besides data privacy should my organization be concerned with?
    \item What does the AIA mean by high-risk AI?
    \item Does the AIA require an external audit?
    \item Which documents should be included in the compliance documentation?
    \item Does the AIA mention metrics that should be used to determine a model’s risks for rights and discrimination?
    \item What does the AIA mean by ‘human oversight’?
\\ \end{enumerate}
Questions on the application of the AIA within their organization:
\begin{enumerate}
    \item To which extent does my ISO certification help towards AIA compliance?
    \item Does GDPR training also include data bias and model bias training?
    \item What are the biggest risks to AIA compliance when data is gathered in-house?
    \item Our organization uses data from customers; what are some of the biggest risks when aiming for AIA compliance?
    \item We only use ChatGPT and other out-of-the-box AImodels; should we still be concerned with the AIA?
    \item What can we do to improve AIA compliance concerning our technical documentation?
    \item We currently don’t communicate anything about our models with our users; how can we better communicate information with the users for AIA compliance?
    \item Our organization is very small, and no one is specialized in compliance; where do we even begin to achieve AIA compliance?
    \item We currently have no idea if we communicate with our stakeholders according to the AIA; how should we assess this to make improvements?
    \item The AIA stipulates that accuracy should be according tothe state of the art. This seems very vague; how should I go about achieving state-of-the-art accuracy?
\end{enumerate}
\end{appendices}

\vspace{12pt}
\end{document}